\documentclass[a4paper,twoside]{article}

\usepackage[utf8]{inputenc}
\usepackage[T1]{fontenc}
\usepackage[english]{babel}
\usepackage{multirow}
\usepackage{amsfonts}
\usepackage{color}
\usepackage{svg}
\usepackage{graphicx} 
\usepackage{csquotes} 
\usepackage{lipsum}
\usepackage[inline]{enumitem}
\usepackage{booktabs}
\usepackage{epsfig}
\usepackage{subcaption}
\usepackage{calc}
\usepackage{amssymb}
\usepackage{amstext}
\usepackage{amsmath}
\usepackage{bbm}
\usepackage{amsthm}
\usepackage{multicol}
\usepackage{pslatex}
\usepackage{apalike}
\usepackage[linesnumbered]{algorithm2e}
\usepackage[bottom]{footmisc}
\usepackage{hyperref}
\usepackage[
  subtle 
]{savetrees}
\usepackage{SCITEPRESS}

\DeclareMathOperator*{\argmax}{arg\,max}
\DeclareMathOperator*{\argmin}{arg\,min}

\begin{document}

\title{Reinforcement Learning Methods for Neighborhood Selection in Local Search}
\author{}
\author{\authorname{
        Yannick Molinghen\sup{1}\orcidAuthor{0009-0008-2757-3737},
        Augustin Delecluse\sup{2,3}\orcidAuthor{0000-0001-6285-6515},
        Renaud De Landtsheer\sup{4}\orcidAuthor{0000-0002-5503-9442} and
        Stefano Michelini\sup{4}\orcidAuthor{0000-0002-8991-9771}}
    \affiliation{\sup{1}Université Libre de Bruxelles, Brussels, Belgium}
    \affiliation{\sup{2}UCLouvain, Louvain-La-Neuve, Belgium}
    \affiliation{\sup{3}KU Leuven, Gent, Belgium}
    \affiliation{\sup{4}CETIC research center, Charleroi, Belgium}
    \email{yannick.molinghen@ulb.be, augustin.delecluse@kuleuven.be,
        renaud.delandtsheer@cetic.be,
        stefano.michelini@cetic.be}
}

\keywords{Local Search, Reinforcement Learning, Combinatorial Optimization}

\abstract{Reinforcement learning has recently gained traction as a means to improve combinatorial optimization methods, yet its effectiveness within local search metaheuristics specifically remains comparatively underexamined. In this study, we evaluate a range of reinforcement learning-based neighborhood selection strategies -- multi-armed bandits (upper confidence bound, $\epsilon$-greedy) and deep reinforcement learning methods (proximal policy optimization, double deep $Q$-network) -- and compare them against multiple baselines across three different problems: the traveling salesman problem, the pickup and delivery problem with time windows, and the car sequencing problem. We show how search-specific characteristics, particularly large variations in cost due to constraint violation penalties, necessitate carefully designed reward functions to provide stable and informative learning signals. Our extensive experiments reveal that algorithm performance varies substantially across problems, although that $\epsilon$-greedy consistently ranks among the best performers. In contrast, the computational overhead of deep reinforcement learning approaches only makes them competitive with a substantially longer runtime. These findings highlight both the promise and the practical limitations of deep reinforcement learning in local search.}

\onecolumn \maketitle \normalsize \setcounter{footnote}{0} \vfill

\section{INTRODUCTION}
Local search (LS) is a widespread paradigm to design metaheuristics for NP-hard optimization problems~\cite{michiels_theoretical_2007,lenstra2018local}. At its core, an LS-based metaheuristic explores neighboring solutions iteratively until a stopping criterion is met. Given a solution to the problem at hand, its set of neighbors is defined through a neighborhood structure, which essentially consists in a set of operators that describe how to alter it. A metaheuristic may use more than one neighborhood to explore the solution space (e.g. variable neighborhood search~\cite{mladenovic1997variable}), and its performance can therefore heavily depend on which neighborhoods are used. Search procedures are typically hand-crafted by domain experts, and may require a substantial amount of tuning to perform well on some instances. Efficient search procedures favor the selection of neighborhood operators quickly leading to high-quality solutions.

Given a set of neighborhood operators to choose from, simple neighborhood selection strategies such as a random or round-robin selection favor diversification. However, these approaches do not exploit the impact of operators on the objective cost to perform meaningful decisions.
In contrast, one can make use of observations from a local search engine such as the runtime of a neighborhood operator or its impact on the objective to design strategies that select the operator with the best expected performance.

Inspired by recent work on multi-armed bandits (MAB) for the selection of heuristics in mixed-integer linear programming solvers~\cite{chmiela23}, we experiment with several strategies for neighborhood selection in LS. Our contributions consist in:
\begin{itemize}
    \item Identifying a list of features and challenges specific to neighborhood selection in LS;
    \item Designing reward schemes based on such features;
    \item Proposing reinforcement learning selection strategies adapted to LS;
    \item Comparing the performance of various selection strategies on three optimization problems, showing that while reinforcement learning selection offer interesting trade-off in performances, they may fail to beat simple baseline strategies.
\end{itemize}
The paper is structured as follows:
Section~\ref{sec:background} covers the background on LS and related work. Section~\ref{sec:methodology} presents the design of reward schemes and reinforcement learning selection considered. The experimental results on their behaviors are outlined in Section~\ref{sec:experiments}.

\section{BACKGROUND}
\label{sec:background}

We start by introducing the key concepts of LS and reinforcement learning, and then discuss related works in the literature.

\subsection{Local Search}
\label{sec:local-search}
We consider combinatorial optimization problems described by a finite set of decision variables with discrete or continuous domains.
Such a problem is represented by the tuple $\langle \mathcal{S}, \mathcal{X}, f\rangle$, where $\mathcal{S}$ is the solution space, i.e., the finite set of all decision variable assignments, $\mathcal{X} \subseteq \mathcal{S}$ is the set of feasible solutions, i.e., solutions for which all constraints necessary to describe the problem are enforced (e.g., vehicle capacity, time windows, etc.), and $f: \mathcal{S} \to \mathbb{R}$ is the objective (or cost) function that measures the fitness of a candidate solution. Without loss of generality, we consider the minimization problem of the form \[
    \text{Find } x^* = \argmin_{x \in \mathcal{X}} f(x).
\]

\paragraph{Move operators and neighborhoods:}
A \emph{move operator} $m$ is a function
\[
    m:\ \mathcal{S}\times\Theta_m \to \mathcal{S},
\]
where a parameter configuration $\theta\in\Theta_m$
specifies how the move is instantiated.
The neighborhood of $s\in\mathcal{S}$ induced by $m$ is
\[
    \mathcal{N}_m(s)=\{\,m(s,\theta)\mid \theta\in \Theta_m\,\}.
\]
A deterministic move is obtained via a selection rule $\sigma$ (e.g., first or best improving solution found in the associated neighborhood), which can then determine the configuration $\theta(\sigma)\in\Theta_m$. Therefore, at iteration $t$ the move $m$ from solution $s_t$ can be expressed as
\[
    s_{t+1}=m\bigl(s_t,\theta(\sigma)\bigr).
\]
In the following, we assume we dispose of a set $\mathcal{M}=\{m_1, m_2,\dots, m_n\}$ of parametrized move operators for each problem under our consideration, and that we make use of such a predetermined selection rule $\sigma_m$ for each $m\in\mathcal{M}$.

\noindent\textit{Example.}
In the traveling salesman problem (TSP), whose solutions are Hamiltonian tours on a graph (see Section~\ref{sec:benchmarks}), the 2-opt~\cite{croes1958method} operator swaps the endpoints $i$ and $j$ of two segments of the tour to obtain a new solution. This operator is thus configured by $\theta=(i,j)$, and by enumerating all possible $(i,j)$ we build the associated 2-opt neighborhood. Given the selection rule $\sigma_m$ we obtain the deterministic move $m(s, \theta(\sigma_m))$ from the current solution $s$ (e.g., find the best-improving $(i,j)$).

\subsection{Reinforcement Learning}
Reinforcement learning (RL) addresses sequential decision-making in which an agent interacts with an environment to maximize its expected return \cite{sutton_reinforcement_2018}.
We evaluate two standard settings that differ by their assumptions about how actions affect the future: the \emph{multi armed bandit} setting where the decision-making does not rely on the current state, and the \emph{Markov decision process} (MDP) setting, where the decision-making is based on the current state and influences future states and rewards.

\paragraph{Bandit setting (single-state):}
The multi-armed bandit (MAB) problem \cite{robbins_some_1952} formalizes the exploration–exploitation tradeoff in a static decision-making context. In an MAB, an agent repeatedly selects one action among a finite set of actions (or arms) $A=\{a_1, \dots, a_n\}$ and receives a stochastic reward according to the pulled arm. Each action $a_i \in A$ yields a stochastic reward $r_t(a_i)$ drawn from an unknown distribution with mean $\mu_{a_i}$.

Equivalently, a bandit can be viewed as a degenerate MDP with a single (unmodeled) state and identity dynamics in which future reward distributions do not change as a function of the agent’s actions. The objective of the agent is to maximize cumulative reward by balancing exploration and exploitation.

\paragraph{Markov decision process (multi-state):}
In the general RL formulation, the environment is modeled as an MDP \cite{bellman_dynamic_1957} $M = \left<S, A, R, T\right>$ where $S$ is the set of states, $A$ is the set of actions, $R: S \times A \times S \to \mathbb{R}$ is the reward function, and $T: S \times A \to \Delta_S$ is the transition function that determines the distribution of landing states $\Delta_S$ given a state-action pair.
The expected return under a given policy $\pi$ is noted \[
    J(\pi) = \mathbb{E}_\pi \left[ \sum_{t=0}^\infty \gamma^t R(s_t, a_t, s_{t+1}) \right]
\]
where $\gamma \in [0, 1]$ is a discount factor parameter that balances the value of immediate rewards in comparison to future ones. The goal of the agent is to find the policy $\pi^\star$ that maximizes $J$, i.e., find $\pi^\star=\argmax_\pi J(\pi)$.

\subsection{Related work}
\label{sec:related-works}

Our approach shares some similarities to the adaptive large neighborhood search (ALNS) introduced by \cite{ropke2006adaptive}, which is based on large neighborhood search \cite{shaw1998using}. Given a list of relaxation and re-optimization procedures, ALNS iteratively selects one such relaxation and re-optimization procedure to improve a solution. Since the performance of such procedures may change during the search, the selection criterion is adapted according to observed performances. However, the work of \cite{turkevs2021meta} has highlighted that the adaptiveness of ALNS only improved the quality of the solution by 0.14\% on average compared to a non-adaptive version where the procedures are chosen randomly. This highlights the need to compare to simple baselines such as the random selection of procedures.

In recent years, RL has been used to guide the search procedure of combinatorial optimization methods. In \cite{cappart2021combining}, a deep RL agent guides the search of a constraint programming solver with a learned heuristic within the branching system of the solver. More recently, the search of a domain-independent dynamic programming has also been guided with reinforcement learning \cite{narita2025reinforcement}. In both works, although a solution is reached in fewer nodes expanded during the search procedure, the resolution time is larger when using RL-based approaches compared to traditional methods due to the computational cost of deep RL.

In the scope of LS specifically, a recent work~\cite{dantas_using_2021} has shown that using deep RL approaches outperform MAB ones when it comes to move operator selection in a vehicle routing problem, even without pre-training and with a tight 5 minutes time budget. Similarly, a deep RL agent is used in \cite{zhang_deep_2022} to guide the search in a container terminal truck routing problem, which yields an improvement in objective function ranging from 3 to 8\% over manual heuristics.

While these studies present good results overall, most of them require an off-line pretraining phase of the RL agent on problem instances comparable to the target one, which requires a dataset of instances of the same size as the target. Moreover, the pretraining duration can be long, taking up to two days in \cite{cappart2021combining}, potentially exceeding the benefit of using RL altogether. Moreover, none of these works compare their results to simple baselines such as random action selection that could potentially outperform their approach. Finally, we argue that these studies do not sufficiently cover the key aspects of agent rewards and state representations, which are necessary to fully leverage the potential capabilities of RL in the scope of combinatorial optimization.

\section{METHODOLOGY}
\label{sec:methodology}

We build on the operators and neighborhoods defined in Section~\ref{sec:local-search}.
At each iteration, the algorithm evaluates the current move operators and decides which one to select. Operators leading to empty neighborhoods are tracked in a tabu list, and the algorithm restarts when all neighborhoods are empty.

\paragraph{Acceptance criterion:}
We only accept improving moves.
Let $\mathcal{M}$ be the set of parameterized move operators and $\Theta_m(s)$ their state-dependent parameters.
The set of admissible improving moves at $s$ is
\begin{multline}
    \label{eq:accept}
    \mathcal{A}(s)=\{\, (m,\theta)\mid m\in\mathcal{M},\ \theta\in\Theta_m(s),\\ f(m(s,\theta))<f(s)\,\}.
\end{multline}
An operator $m$ is \emph{improving} at $s$ if there exists $\theta\in\Theta_m(s)$ with $(m,\theta)\in\mathcal{A}(s)$; otherwise it cannot yield an improving move at $s$.
When attempting to apply a move $(m, \theta)$, the solution $s_t$ is only updated if the move is admissible, and is kept otherwise:
\begin{equation}
    \label{eq:new-solution}
    s_{t+1}=\begin{cases}
        m(s_t,\theta) & \text{ if } (m,\theta)\in\mathcal{A}(s_t) \\
        s_t           & \text{ otherwise }
    \end{cases}
\end{equation}

\paragraph{Constraints:}
Constraints are treated in two ways.
\emph{Hard constraints} are enforced by restricting the neighborhood to moves that preserve feasibility (i.e., $m(s,\theta)\in\mathcal{X}$ whenever $s\in\mathcal{X}$).
\emph{Soft constraints} may be violated, in which case the objective is decomposed as
\begin{equation}
    \label{eq:decomp}
    f(s)=c(s)+v(s),
\end{equation}
where $c:\mathcal{S}\to\mathbb{R}$ is the cost term and $v:\mathcal{S}\to\mathbb{R}^+$ measures violation, with $v(s)=0$ iff $s\in\mathcal{X}$.
We assume feasibility strictly dominates infeasibility, i.e.,
\begin{equation}
    \label{eq:dominance}
    \forall x\in\mathcal{X},\ \forall s\in\mathcal{S}\setminus\mathcal{X}:\quad f(x)<f(s).
\end{equation}
This can for instance be achieved by scaling $v$ so that $\max_{x\in\mathcal{X}} c(x) < \min_{s\notin\mathcal{X}} v(s)$. With such formulation, the image domain of $f$ is likely to span across multiple orders of magnitude, and we do not require to start from an initial feasible solution but only from a solution satisfying the hard constraints.

\paragraph{Tabu list:}
Evaluating a move operator to determine whether the associated neighborhood contains an improving solution carries a certain computational cost. Thus, to prevent the selection of move operators that have already been shown not lead to improving solutions, we maintain at step $t$ a tabu list $\mathcal{T}_t\subseteq\mathcal{M}$ of operators temporarily not admissible for selection.
The admissible improving moves are
\begin{multline}
    \label{eq:admissible-moves}
    \mathcal{A}_t(s)=\{\, (m,\theta)\mid m\in\mathcal{M}\setminus \mathcal{T}_t,\ \theta\in\Theta_m(s),\\ f(m(s,\theta))<f(s)\,\}.
\end{multline}
We select and apply some $(m,\theta)\in\mathcal{A}_t(s)$, yielding $s_{t+1}$ according to Equation~\eqref{eq:new-solution}. Operators that have been attempted but found non-improving (i.e., $f(m(s_t,\theta))\ge f(s_t)$) are added to $\mathcal{T}_t$. After a move is accepted, the tabu list is reset: $\mathcal{T}_{t+1}=\emptyset$.

\paragraph{Restarts:}
It is often the case that LS methods may reach a local minimum $s\in\mathcal{S}$, i.e., a solution for which the improving set is empty (or there is no admissible improvement under $\mathcal{T}_t$).
When this occurs, we perform a \emph{restart} to escape from the local minimum, e.g., by sampling a new candidate in $\mathcal{S}$.

\subsection{Reinforcement Learning Formulation}
We model the LS procedure as a fully observable single-agent RL problem. Given an initial solution $s_0 \in \mathcal{S}$, the goal consists in finding the sequence of moves that minimizes the objective function $f$ in the fewest steps possible. Formally, we model the procedure as an MDP $M=\left<S, A, R, T\right>$ where:

\begin{itemize}
    \item $S$ is the state space, corresponding to the solution space $\mathcal{S}$.
    \item $A = \left\{a_1, a_2, \dots, a_n\right\}$ is the set of actions.
          Formally, an action is a deterministic map $a:\mathcal{S}\to\mathcal{S}$. In this study, we associate each available move operator to an action \[
              a_i(s)=m_i\bigl(s,\theta(\sigma_{m_i})\bigr)\quad \forall i \in \{1,2, \dots,n\}.
          \]
    \item $R: S \times A \times S \to \mathbb{R}$ is a function that determines the reward for taking action $a$ in state $s_t$ and landing in $s_{t+1}$. We describe in Section~\ref{sec:reward-function} the reward functions that we consider in this work.
    \item $T: S \times A \to S$ is a deterministic transition function that maps each state-action $(s_t, a)$ pair to its next state $s_{t+1}$. This transition is made by applying the selected action to the given state, i.e., applying the move to the solution.
\end{itemize}

An important concept in RL is the \emph{episode}, which is a complete sequence of states and actions from a \emph{start} state to a \emph{terminal} state from which no further actions can be taken, and therefore no further reward can be collected. In the scope of this work, a terminal state is thus a state from which a restart must be performed, i.e., a local minimum. The global LS--RL loop is shown in Algorithm~\ref{algo:rl-loop}.

\begin{algorithm}
    \DontPrintSemicolon
    \caption{LS--RL loop.}
    \label{algo:rl-loop}
    \KwIn{$M=\left<S, A, R, T\right>$, \textit{agent}: the RL agent}
    $s_0 \sim \text{Uniform}(S)$\;
    $t\gets 0$\;
    $\mathcal{T}_t \gets \emptyset$\;
    \While{time remaining}{
        $m \gets $\textit{agent}.get\_move($s_t, \mathcal{T}_t$)\;
        $a \gets (m, \theta(\sigma_m))$\;
        \eIf{$a \in \mathcal{A}_t$}{
            $\mathcal{T}_t \gets \emptyset$ \;
            $s_{t+1} \gets m(s_t, \theta)$
        } {
            $\mathcal{T}_t \gets \mathcal{T}_t \cup \{m\}$\;
            \eIf{$|\mathcal{T}_t| = |A|$}{
                \textit{agent}.terminate\_episode()\;
                $s_{t+1} \sim \text{Uniform}(S)$
            }{
                $s_{t+1} \gets s_t$\;
            }
        }
        $r \gets R(s_t, a, s_{t+1})$\;
        \textit{agent}.learn($s_t, a_m, r, s_{t+1}$)\; \label{line:learn}
        $t\gets t + 1$\;
    }
\end{algorithm}

\subsection{Bandit Algorithms}
\label{sec:bandits}

Our first contribution resides in the exploration of two MABs to guide the selection of the move operator at each step of the search: $\epsilon$-greedy and upper confidence bound (UCB). Over the course of the search, an MAB learns an estimate of the instantaneous reward that each action yields. To do so, it maintains a vector $\hat{r} = \left<\hat{r}_{a_1}, \hat{r}_{a_2}, \dots, \hat{r}_{a_n}\right>$. At each time step $t$, after receiving the reward $r_t$ for taking action $a_i$, the agent updates the reward estimate in the following way\[
    \hat{r}_{a_i,t+1} = (1-\alpha)\hat{r}_{a_i,t} + \alpha r_t,
\]
where $\alpha \in (0, 1]$ is a learning rate. This update is performed at line~\ref{line:learn} of Algorithm~\ref{algo:rl-loop}. Each MAB differs in its approach to the exploration-exploitation tradeoff and in the action selection mechanism.

    \paragraph{$\epsilon$-greedy:} The $\epsilon$-greedy algorithm balances exploration and exploitation according to a parameter $\epsilon \in [0, 1]$. This MAB takes the action with the highest expected reward with probability $1-\epsilon$ and a random action otherwise. Given a randomly sampled value $p \sim \text{Uniform}([0, 1))$ at each time step $t$, the action selection criterion is:
\[
    a_t = \begin{cases}
        \argmax_{a \in \mathcal{A}_t(s_t)} \hat{r}_a & \text{ if } p > \epsilon, \\
        \sim \text{Uniform}( \mathcal{A}_t(s_t))     & \text{ otherwise.}
    \end{cases}
\]
Note that $\epsilon=0$ results in a fully greedy algorithm and $\epsilon=1$ in a fully random one.

\paragraph{Upper confidence bound:} The UCB algorithm approaches the exploration–exploitation trade-off by maintaining an upper estimate of each action's potential and taking the action with the highest estimate. To do so, each arm is associated with the number of times that action has already been taken ($n_a$). UCB first selects each action once, then selects the best action from $t > n$ as \[
    a_t = \argmax_{a \in \mathcal{A}_t(s_t)} \hat{r}_a + c \sqrt{\frac{2 \ln t}{n_a}},
\]
where $c \in \mathbb{R}^+$ is a confidence parameter.

\subsection{Deep Reinforcement Learning}
\label{sec:deep-rl}
Our second contribution lies in the evaluation of two deep RL (DRL) algorithms applied to LS: double deep $Q$-network \cite[DDQN]{van_hasselt_deep_2016} and proximal policy optimization \cite[PPO]{schulman_proximal_2017}. Because the size of the search space $S$ is very large, tabular methods such as $Q$-learning~\cite{watkins_q-learning_1992} are not applicable because they would either not fit in memory, or be extremely slow to converge.

One of the main questions of DRL in the scope of LS is how to feed the current state of the search to the neural network (NN) of the DRL agent. Since we study three different problems (Section~\ref{sec:benchmarks}) with different structures, we need different types of NN. In particular, we leverage graph neural networks \cite[GNN]{kipf_semi-supervised_2017} for routing problems and convolutional neural networks \cite[CNN]{lecun_gradient-based_1998} in scheduling ones to take advantage the locality of the information. In both cases, the NN has one output neuron per action. A complete description of the state encoding is detailed for each problem in Section~\ref{sec:benchmarks}, and we present the corresponding GNN and CNN architectures in Table~\ref{tab:gnn-architecture}.

\begin{table}[ht]
    \setlength\tabcolsep{0pt}
    \caption{NN architectures for DRL.}
    \small
    \centering
    \label{tab:gnn-architecture}
    \begin{tabular}{lr}
        \toprule
        \multicolumn{2}{c}{\textbf{GNN}}                                 \\
        Layer               & Output shape                               \\
        \midrule
        Input               & ($V$: \#vertex attrs., $E$: \#edge attrs.) \\
        GCNConv             & ($V$: 32, $E$: 3)                          \\
        Leaky ReLU          & ($V$: 32, $E$: 3)                          \\
        GCNConv             & ($V$: 64, $E$: 3)                          \\
        Global Max Pool     & 64                                         \\
        Linear              & 64                                         \\
        Leaky ReLU          & 64                                         \\
        Linear              & $n$                                        \\
        \midrule
        \multicolumn{2}{c}{\textbf{CNN}}                                 \\
        Layer               & Output shape                               \\
        \midrule
        Input               & Matrix: $m\times n$                        \\
        Conv1D + Leaky ReLU & $32 \times (n-2)$                          \\
        Conv1D + Leaky ReLU & $64 \times (n-4)$                          \\
        Conv1D + Leaky ReLU & $16 \times (n-6)$                          \\
        Flatten             & $x = 16 \times (n - 6)$                    \\
        Linear + Leaky ReLU & $x/ 2$ until $x/2 \leq 192$                \\
        Linear              & $n$                                        \\
        \bottomrule
    \end{tabular}
\end{table}

Note that unlike some other works in the field, we do not train our DRL agents in advance, which would require a dataset of instances whose state shape is compatible with the instance at hand. Rather, we train our DRL agents from scratch on each instance as shown at line~\ref{line:learn} of Algorithm~\ref{algo:rl-loop}.

\paragraph{DDQN:} DDQN is a value-based DRL algorithm that relies on the learning of a $Q$-function that associates a value to each state-action pair. At each step, we take an action according to an $\epsilon$-greedy algorithm
\[
    a_t = \begin{cases}
        \argmax_{a \in \mathcal{A}_t(s_t)} Q(s_t, a) \quad & \text{if }p\geq \epsilon_{DDQN} \\
        \sim \text{Uniform}(\mathcal{A}_t(s))              & \text{otherwise.}
    \end{cases}
\]

\paragraph{PPO:} PPO is an actor-critic policy-gradient algorithm that directly outputs a probability distribution $\Delta_{\mathcal{A}_t(s)}$ over the (non-tabu) actions $\mathcal{A}_t(s)$. The action is thus directly sampled from this distribution
\[
    a_t \sim \Delta_{\mathcal{A}_t(s)}.
\]

\subsection{Reward Functions}
\label{sec:reward-function}
An important feature of the reward function is that it must be aligned with the objective of the LS solver, i.e.,the minimization of $f$. In this section, we discuss and justify our approach to that end.

\subsubsection{Theoretically Ideal Reward Function}
\label{sec:ideal-reward}
We make the connection between $V: S \to \mathbb{R}$, the value function from the RL point of view, and $f: \mathcal{S} \to \mathbb{R}$ the objective function from the LS point of view. Since RL formalizes the decision-making problem as a maximization problem, and the LS solver as a minimization one, we can define the value of a state $V(s)$ as the opposite of its cost $f(s)$ \[
    V(s) = -f(s).
\]
Then, we leverage potential-based reward shaping~\cite[PBRS]{y_ng_policy_1999} to design a reward function guaranteed to align with the value function. PBRS defines the shaped reward function as \[
    R(s_t, a, s_{t+1}) = R_i(s_t, a, s_{t+1}) + \gamma \,\phi(s_{t+1}) - \phi(s_t),
\]
where $R_i$ is the initial reward function. To design our reward function, we choose $\phi=V$ (as suggested by the original authors), an \emph{empty} initial reward function, i.e.,$R_i(s_t, a, s_{t+1}) = 0$ for all $ s_t, a, s_{t+1}$, and $\gamma=1$. The resulting reward function $R_1$ is defined as follows:
\begin{align}
    \label{eq:ideal-reward}
    \begin{split}
        R_1(s_t, a, s_{t+1}) & = V(s_{t+1}) - V(s_t) \\
                             & = f(s_t) - f(s_{t+1})
    \end{split}
\end{align}
Intuitively, the reward from $s_t$ to $s_{t+1}$ is the improvement in cost. In particular, the theoretical guarantees of PBRS ensure that the optimal policy $\pi^\star$ is aligned with the objective, i.e., the maximization of $V$ and therefore the minimization of $f$.

\subsubsection{Log-Adjusted Reward Function}
Even though the reward function of \eqref{eq:ideal-reward} is ideal in theory, for problems whose objective function has the form $f(s) = c(s) + v(s)$ (see \eqref{eq:decomp}), $R$ spans several orders of magnitude. Actions that remove hard constraint violations will yield rewards that are orders of magnitude higher than actions that improve a feasible solution with respect to its cost.

As a result, in both DRL algorithms, the gradients of state-actions pairs that remove a hard constraint will largely dominate the ones of other actions. Overall, the DRL optimization process ends up in a local minimum where either large or small values are estimated accurately, but not both. To mitigate this effect, we work with a reward function $R_2$ in the log space \[
    R_2(s_t, a, s_{t+1}) = \max{\left(0, \log_{10}(V(s_{t+1}) - V(s_t))\right)}.
\]
A direct consequence of such reward function for RL agents is that it is more beneficial to perform three actions yielding a cost improvement of 10 (i.e., a reward of $3 \times \log_{10}(10)=3$ and a global cost improvement of 30) than a single action with a cost improvement of 100 (i.e., a reward of $\log_{10}(100)=2$. However, using such a reward scheme does not have the same implications for MABs, as they only rely on immediate rewards and do not consider long-term return.

\subsubsection{Duration-Aware Reward Function}
In order to incentivize the agent to select move operators that rapidly yield a substantial improvement, we design a third reward function that takes into account the following criteria:
\begin{enumerate}
    \item whether the move yielded an improvement or not;
    \item the move elapsed computation time $E(s_t, a, s_{t+1})$;
    \item the slope of the objective gain yielded by the move with regard to the computation duration.
\end{enumerate}
Therefore, we define this reward function as follows:
\begin{align}
    R_3(s_t, a, s_{t+1}) = & w_1 \times \mathbbm{1}(f(s_{t+1}) < f(s_t))                  \\
    +                      & w_2 \times E(s_t, a, s_{t+1})                                \\
    +                      & w_3 \times \frac{f(s_{t}) - f(s_{t+1})}{E(s_t, a, s_{t+1})},
\end{align}
where $w_1, w_2$ and $w_3$ are hyperparameters associated with the aforementioned criteria, and $\mathbbm{1}$ is the indicator function.

\section{COMPUTATIONAL EXPERIMENTS}
\label{sec:experiments}
We present the combinatorial optimization problems on which we perform our experiments, describe our experimental scenario and finally discuss our results.

\subsection{Considered Problems}
\label{sec:benchmarks}
We consider the three following problems: the TSP, the Pickup and Delivery Problem with Time Windows (PDPTW) and the Car Sequencing Problem (CSP).

\subsubsection{Traveling Salesman Problem}

The TSP~\cite{flood1948tsp} is a combinatorial optimization problem that, given a set of cities, consists in finding the shortest possible Hamiltonian tour, i.e., the shortest possible tour that visits each city exactly once.
Formally, for an instance with $n$ cities and a distance matrix $D$ of size $n\times n$ where $D_{i,j}$ is the distance between cities $i$ and $j$, the objective is to determine the path of cities $p = \{p_1, p_2, \dots, p_n, p_1\}$ where $p_i \neq p_j$ if and only if $i\neq j$, that minimizes the total travel distance $f(p)$
\begin{equation}
    \label{eq:tsp}
    f(p) = \sum_{i=1}^{n} D_{p_i, p_{i+1}}
\end{equation}
where $p_i$ denotes the $i$\textsuperscript{th} city of the path $p$ represented in the solution $s$.

This problem is modeled using one sequence variable~\cite{de2018reasoning}, capturing the travel path. It starts empty and may be grown using an \texttt{insert} operator, adding an unrouted node to the path. Two other operators, \texttt{move} and \texttt{2-opt}, either move a node or perform a 2-opt operation \cite{croes1958method} to alter the order of visits of the nodes. The objective function corresponds to the total travel distance~\eqref{eq:tsp} and a constraint penalty for unrouted nodes.
This formulation is not the most effective one to solve TSP with LS, but let us work with an objective including a penalty term on a relatively simple optimization problem.

\begin{table*}[t]
    \centering
    \caption{Summary of reward schemes and their uses across the three problems under study.}
    \begin{tabular}{lp{4.5cm}p{6cm}l}
        \toprule
        \textbf{Reward} & \textbf{Description}                                       & \textbf{Characteristics}                   & \textbf{Used in} \\
        \midrule
        $R_1$           & Objective gain                                             & Theoretically aligned with the LS process. & CSP              \\
        $R_2$           & Log objective gain                                         & Less prone to penalty dominance.           & TSP, PDPTW       \\
        $R_3$           & Weighted sum of objective gain, computation time and slope & Takes computation time into account.       & TSP, PDPTW       \\
        \bottomrule
    \end{tabular}
    \label{tab:reward-summary}
\end{table*}

\paragraph{DRL State representation:}
A solution to the TSP can be represented as a graph where the vertices are the cities (i.e.,the problem statement) and the edges are the ones that constitute the tour. The attributes of each vertex are its normalized coordinates ($\in [0, 1]$) and a distance vector to every other vertex, while the attributes of each edge $e_{i,j}$ are the time of departure from vertex $i$ and the time of arrival to vertex $j$.

\subsubsection{Pickup and Delivery Problem with Time Windows}
The PDPTW~\cite{dumas1991_pdptw} is a Vehicle Routing Problem in which a set of transportation requests must be served by one or several vehicles. Each request consists of a pickup location and a corresponding delivery location, and both must be visited by the same vehicle in the correct order—pickup before delivery. Additionally, each location $i$ is associated with a time window $\left[e_i,l_i\right]$, specifying the earliest and latest times at which service can occur.
The goal is to determine a set $s$ of vehicle routes that satisfies all requests, respects each vehicle's maximal capacity and each location's time window constraint, and minimizes the total travel length across all vehicles.
Formally, the cost function is \[
    f(s) = \sum_{v \in V} \sum_{i \in N} \sum_{j \in N} c_{i,j} \cdot x^v_{i,j}
\]
where $V$ is the set of vehicles, $N$ is the set of pickup and delivery locations (including depots), $c_{i,j}$ is the travel path length from location $i$ to location $j$, and $x^v_{i,j}$ is a binary variable equal to 1 if vehicle $v$ travels directly from $i$ to $j$, and 0 otherwise.

Similarly to the TSP, this problem is modeled using sequence variables. The violations of time windows, capacity of vehicles and precedences between pickups and deliveries are treated as constraint penalty and added to the objective term as described in~\eqref{eq:decomp}. The move operators consists in inserting an unrouted pair of pickup and delivery, moving one node or a pair of pickup and delivery, and exchanging two segments of two paths.

\paragraph{DRL State representation:} Similarly to the TSP, we represent the PDPTW as a graph. In this problem, in addition to the vertex attributes already stated for the TSP, each vertex is assigned an opening and a closing time, a service duration, an attribute $\Delta_{\text{load}}$ that indicates the difference in vehicle load ($\Delta_{\text{load}}> 0$ for pickups and $\Delta_{\text{load}}< 0$ for deliveries), and a unique delivery identifier. Similarly, the additional attribute for the edges is the vehicle load during the travel.

\begin{figure*}[!ht]
    \centering
    \includegraphics[width=\linewidth]{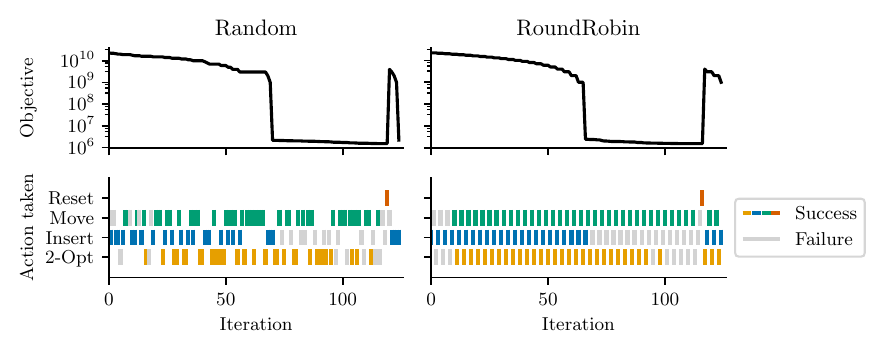}
    \caption{First stages of the search for the random and RR move selectors on a small TSP instance (24 cities). The top plot shows the value of $f$ over the course of the search, and the bottom plot shows which move operator has been tried but not selected because they did not improve the solution (grayed), and the ones that have been selected (colored) at each iteration.}
    \label{fig:tsp-action-baselines}
\end{figure*}

\subsubsection{Car Sequencing Problem}
The CSP~\cite{parrello1986csp} is a combinatorial optimization problem arising in automotive manufacturing, where a sequence of cars must be arranged on an assembly line to satisfy both production capacity and option constraints. Each car belongs to a given class that determines which optional features (such as sunroofs, air conditioning, or specific engines) it requires. The assembly line has limited capacity for installing each option—typically expressed as a ratio $p/q$, meaning that at most $p$ cars requiring that option can appear within any subsequence of $q$ consecutive vehicles. The goal is to determine a feasible production sequence that satisfies these ratio constraints. Formally, let $n$ be the number of cars to sequence, $O$ the set of options, and $a_{i,o} \in \left\{0, 1 \right\}$ a binary parameter indicating whether car $i$ requires option $o$. If $s=(s_1, s_2, \dots, s_n)$ denotes the car sequence, and $p_o/q_o$ is the constraint ratio of option $o \in O$, the objective function can be expressed as minimizing the total number of constraint violations as: \begin{equation}
    \label{eq:constr_viol}
    f(s) = \sum_{o \in O} \sum_{t=1}^{n-q_o - 1} \max \left(0, - p_o +\sum_{i=t}^{t + q_o - 1} a_{s_i, o} \right)
\end{equation}
where the $\max()$ component of Equation~\eqref{eq:constr_viol} counts the number of violated constraints.

This problem is modeled using an array of integer variables, describing the car sequence being produced.
The neighborhoods consist in swapping two cars within the sequence, moving one car to another position, flipping one subsequence of cars, or swapping two subsequences of cars.

\paragraph{DRL State representation:} We represent the state of a CSP as two matrices. The first matrix encodes the state of the search and is a matrix of size $|O| \times n$ where the $i$\textsuperscript{th} column is the one-hot encoding of the options required to produce a specific car at step $i$. The second matrix of size $|O|\times 2$ encodes the maximal utilization ratio $p/q$ of each option.

\subsection{Experimental Setup}
\label{sec:experimental-setting}
We evaluate our methods and reward functions on the three problems previously presented in Section~\ref{sec:benchmarks}. Table \ref{tab:reward-summary} summarizes which rewards are considered on which problem. We use 80 TSP instances from TSPLib~\cite{reinelt1991tsplib} with 14 to 1000 cities, 82 CSP instances from CSPLib~\cite{gent1999csplib} with sequence lengths from 100 to 500, and 206  PDPTW instances from~\cite{lilim} with 25 to 150 vehicles and 100 to 600 deliveries. For each problem, we reserve 50\% of the instances for the hyperparameter tuning with \texttt{irace} \cite{irace}, and the remaining instances are left for the final evaluation between the methods.
The tuned parameters are the weight coefficients of the $R_3$ reward function as well as the agent-specific parameters (e.g., $\epsilon$ for $\epsilon$-greedy). The tuning of DDQN with $R_1$ on the TSP and the PDPTW converged to random policies with $\epsilon_{DDQN}\approx 1$, thus we did not incorporate them.

As we aim to compare RL methods across a wide range of problems, we rely on variables and move operators that are readily extensible to new problem variants. The best-performing move for each problem was deliberately excluded, and the best solutions we obtain are not as good as state-of-the-art results. We conjecture that this design better captures deployment on previously unseen problems, where efficient move operators may not yet have been identified.

We compare our MAB and RL methods against three baseline move selectors:
\begin{itemize}
    \item Random, which uniformly samples among available moves;
    \item Round-robin (RR), which returns the next available move in a predetermined cycle;
    \item Best slope first (BSF), which always selects the move with the highest slope, averaged over previous calls.
\end{itemize}
All experiments were conducted using AMD EPYC 7763 CPU in single-threaded mode and an NVIDIA A100 GPU. The implementation of the LS components was done in Scala using the OscaR-CBLS constraint-based local search solver~\cite{oscar_2013}. The MABs were implemented in Scala, while the DRL algorithms were implemented in Python with a Unix socket interconnection layer with the OscaR solver.

\subsection{Insights From a Small TSP Instance}

Figure~\ref{fig:tsp-action-baselines} shows the behavior of two move selection baselines -- random and RR -- on a small TSP instance with 24 nodes. The top part shows the evolution of $f$, and the bottom one the move operator selection and whether its application found an empty neighborhood, in which case it is considered as a failure. Both baselines obviously fail to detect when some operators always lead to empty neighborhoods, and select \texttt{move} and \texttt{2-opt} operators when few nodes are routed (very early iterations), or select \texttt{insert} although all nodes are routed.

\begin{figure*}[!ht]
    \centering
    \includegraphics{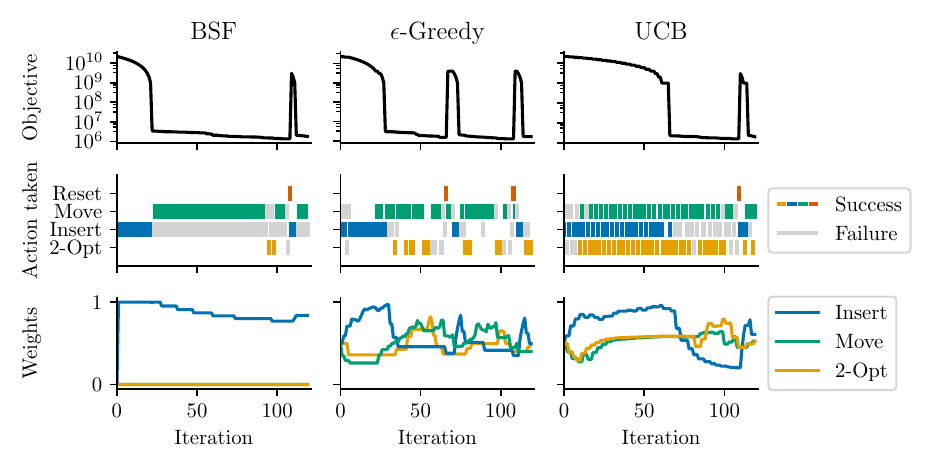}
    \caption{BSF, $\epsilon$-greedy and UCB move operator selection on the TSP. The weights are normalized between [0, 1].}
    \label{fig:tsp-action-egreedy-ucb}
\end{figure*}

In contrast, Figure~\ref{fig:tsp-action-egreedy-ucb} shows the behavior of BSF, $\epsilon$-greedy and UCB with their respective move selection and weights evolution of the operators is also shown at the bottom.
Given the large impact of constraint penalty (i.e.,the unrouted nodes) on the objective function~\eqref{eq:dominance}, BSF always tries to apply an \texttt{insert} move as soon as it is not tabu, as the other moves do not impact the penalty.
By doing so, it correctly identifies that the \texttt{insert} operator is relevant at the beginning of the search and when nodes are unrouted.
However, it only selects the other operators once all nodes are routed, although using a few \texttt{move} and \texttt{2-opt} operators on partially completed paths may already slightly improve the solution.
This justifies the normalization choices explained in Section \ref{sec:reward-function}.

Regarding the bandit agents, $\epsilon$-greedy intertwines \texttt{move} and \texttt{insert} operators even when the path does not visit all nodes, and \texttt{2-opt} move are chosen more often.
Finally, UCB diversifies more the selection but takes longer to favor \texttt{move} and \texttt{2-opt} operators when all nodes are routed, due to the slow weight decrease for the \texttt{insert} operator.

\subsection{Results}
\label{sec:results}
We presents our results with regard to the primal gap $g(s)$ of a solution $s\in\mathcal{S}$ with respect to a given instance of each considered problem~\cite{berthold2013measuring}:
\begin{equation}
    \label{eq:primal-gap}
    g(s) = \begin{cases}
        0                                            & \text{if } f(s) = c^* \\
        1                                            & \text{if } v(s) > 0   \\
        \frac{|c(s) - c^*|}{\max{\{|c^*|, |c(s)|\}}} & \text{otherwise.}     \\
    \end{cases}
\end{equation}
The primal gap $g(s)$ measures the gap between the cost $c^*$ of the best known solution (reported in ~\cite{gent1999csplib,reinelt1991tsplib,lilim}) and the cost $c(s)$ of the solution $s$ in the form of \eqref{eq:decomp}. A solution with a low primal gap is therefore a solution of high quality. This metric allows to compare results across instances of different sizes and to account for infeasibility. In our results, we plot the primal gap of the best solution found over the course of the search.

\begin{figure*}[!ht]
    \centering
    \includegraphics[width=\linewidth]{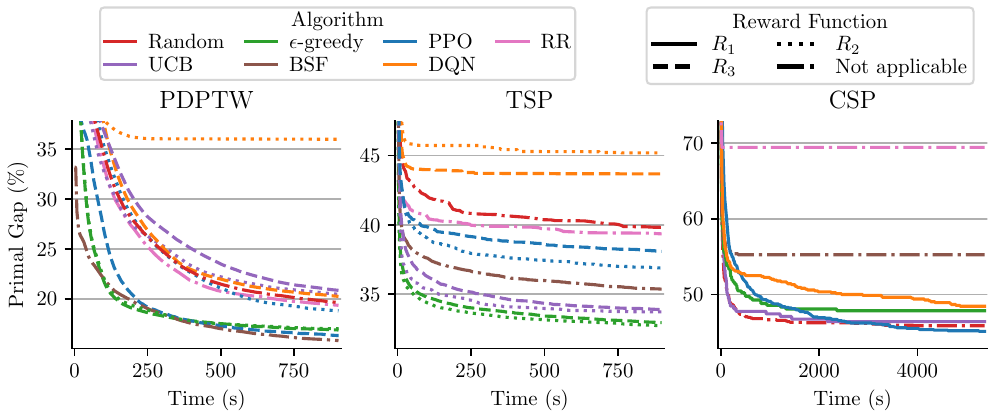}
    \caption{Primal gap (lower is better) over time averaged across all instances and 10 random seeds. Colors account for the move selection algorithm and the line styles for the reward function. The time budget is 15 minutes for the PDPTW and the TSP, and 90 minutes for the CSP. The CSP is the only problem where DRL agents perform the best.}
    \label{fig:subplot-gap-over-time}
\end{figure*}

Figure~\ref{fig:subplot-gap-over-time} shows our results for each method on the TSP, CSP, and PDPTW, expressed as the average primal gap over 10 different random seeds. We set the time limit to 15 minutes for the TSP and PDPTW and to 90 minutes for the CSP, as DRL approaches exhibited a slower convergence on this problem. The performance of each method varies greatly depending on the problem, reward scheme and time budget. Overall, the best move selector varies substantially depending on the problem, and it is clear that there is no single best performer across all problems. Yet, $\epsilon$-greedy stands out as the most reliable algorithm by consistently ranking among the best performers.

\subsubsection{Problem-Wise Analysis}
On the PDPTW, BSF is the best performer with an average primal gap around 15\%, with PPO and $\epsilon$-greedy following closely. We can also observe that DDQN is clearly suboptimal with $R_2$, and that no further improvement is likely to occur since its curve flattens before the end of the time budget.

On the TSP, there is a clearer ordering of the methods regardless of the reward function, as the reward variants of most methods are close to each other performance-wise. In this benchmark, $\epsilon$-greedy is the best performer (with $R_2$, then $R_3$). We note that DDQN stalls with both reward functions, resulting in a flattened curve for most of the search. Additionally, the fact that BSF outperforms both the Random and RR baselines suggests that intensification is more effective than diversification in this case.

The CSP is the only problem where PPO is the best performer, although it only surpasses the random selector -- the second best performer -- after one hour. Additionally, let us note that both DRL methods keep improving until the end, which suggests that a larger time budget may yield even better performance. In contrast to the TSP, BSF falls behind random on this problem, suggesting that diversification is more important than intensification in that scenario.

\subsubsection{Reward-Wise Analysis}

Our results on the PDPTW and the TSP indicate that both MABs consistently perform better (or equally) with the $R_2$ reward scheme (in the $\log$ space) than with the $R_3$ one which aims at incentivizing rapid improvements. This is however not the case for DRL methods, where no clear ordering of the reward functions can be observed. Still, we note that PPO outperforms all other methods on the CSP that uses the $R_1$ scheme, i.e., the \emph{ideal} reward function discussed in Section~\ref{sec:ideal-reward}. On the PDPTW, we note that DDQN performs on par with most algorithms with $R_3$ but perform the worst with $R_2$, where it stalls around 33\%.

In general, we observe that PPO is less sensitive to the reward function than DDQN, in the sense that the results of the former with the two reward functions are closer to each other than the latter. We attribute this difference to the fact that DDQN is a value-based DRL method, meaning that the $Q$-function directly trained on the reward signal impacts the action selection.  In contrast, PPO is an actor-critic method where the learned value function is only used to guide the actor network during training but not to select actions.

\subsubsection{Time-Wise Analysis}
It is important to note that in most cases, the search does not stabilize by the end of the time budget, suggesting that the performance ordering could vary if more time were allocated. This is particularly the case for DDQN (and for PPO to a lesser extent) in the CSP, which has a more pronounced slope than the other algorithms by the end of the time budget.

It is expected that DRL algorithms may still improve later in the process, since they are trained on the fly and require some time to converge. Moreover, DRL methods interact less frequently with the LS solver than that their counterparts because of their superior computational overhead, which is caused by the Scala-to-Python interconnection, as well as by the NN inference and backpropagation.

\section{CONCLUSIONS}
We have studied the use of RL strategies -- covering both MABs and DRL methods -- for neighborhood selection in LS. We have identified search-specific challenges such as highly non-uniform cost landscapes arising from constraint violation penalties and the need for reward signals to be aligned with the LS optimization process. Based on these observations, we have designed three problem-agnostic reward schemes including one theoretically grounded in potential-based reward shaping that rewards the agent proportionally to the cost improvement.

Our experiments with fine-tuned algorithms across three different combinatorial optimization problems -- the TSP, the PDPTW and the CSP -- reveal that no single method dominates in all settings, although $\epsilon$-greedy consistently ranked among the best performing methods. While capable of achieving competitive or even superior performance on the CSP, DRL methods were found to be significantly constrained by their computational overhead, interacting with the solver less frequently than other methods. Their advantages materialize only with substantially longer time budgets, and their sensitivity to reward scale -- especially in the DDQN case -- emphasizes the importance of carefully designed reward functions. PPO proved to be more stable than DDQN in all circumstances, suggesting that actor-critic methods may be better suited for neighborhood selection in LS, although the reasons for this require further investigation.

Overall, our results highlight both the promise and the limitations of RL-based neighborhood selection. While DRL methods can automatically adapt operator usage to instance-specific structures, their benefits are far from guaranteed and must be weighed against their computational cost and the complexity of designing appropriate reward signals. In particular, our findings indicate that simple and inexpensive methods such as $\epsilon$-greedy remain remarkably strong baselines thanks to their balance between adaptivity and efficiency, and should therefore not be overlooked when evaluating advanced learning-based heuristics. These findings suggest that RL should not be viewed as a direct replacement for classical LS heuristics, but rather as a tool whose practical relevance must be carefully assessed against strong, low-overhead baselines and realistic time constraints.


Several directions emerge for future work. First, it remains to be investigated whether the conclusions of this study generalize to other LS solvers, alternative neighborhood structures, or additional classes of optimization problems. Then, the role of pretraining requires further study. While offline pretraining could improve the early performance of DRL methods, it introduces additional costs and assumptions on instance similarity and DRL generalization capability. A comparative study between online, offline and hybrid training schemes would be valuable to better understand the trade-offs between computational overhead and solution quality. Ultimately, this work calls for a more critical and context-aware integration of RL into local search where simplicity and efficiency remain decisive factors.

\section*{ACKNOWLEDGMENTS}
This work was supported by Service Public de Wallonie Recherche under grant n°2010235 -- ARIAC by DIGITALWALLONIA4. The present research benefited from computational resources made available on Lucia, the Tier-1 supercomputer of the Walloon Region, infrastructure funded by the Walloon Region under the grant agreement n°1910247. We would also like to thank the members of the Combinatorial Algorithmics group at CETIC (Thomas Fayolle, Fabian Germeau and Gustavo Ospina) for their support on this project.

\bibliographystyle{apalike}
{\small
    \bibliography{references}}

\section*{\uppercase{Appendix}}

\subsection*{Parameter Tuning}
\label{apx:tuning}

The tuning was performed using \texttt{irace} \cite{irace} by running a given configuration for 5 minutes (30 for the CSP) on a training instance and keeping the configurations minimizing the integral of the primal gap \eqref{eq:primal-gap} \cite{berthold2013measuring}.
For PPO, we tune the $c_1$ and $c_2$ (i.e., actor and critic loss weights) start and end value, the annealing duration, the batch and minibatch sizes as well as the number of training epochs, and the actor and critic learning rates. For DDQN, we tune the exploration parameter $\epsilon_{DDQN}$, the replay memory size, the batch size and the learning rate.
The weights of the $R_3$ reward were only tuned for $\epsilon$-greedy and UCB.
The values identified with a budget of 800 runs for each tuned agent and reward are presented in Tables~\ref{tab:parameters-r3}, \ref{tab:parameters-ppo}, \ref{tab:parameters-ddqn} and~\ref{tab:paramters-egreedy-ucb}.

\begin{table}[ht]
    \caption{$R_3$ weights.}
    \label{tab:parameters-r3}
    \centering
    \small 
    \bgroup
    \begin{tabular}{llrrc}
        \toprule
        Problem                & $w$   & $\epsilon$-greedy & UCB   & DDQN \& PPO \\
        \midrule
        \multirow{3}{*}{TSP}   & $w_1$ & 0.614             & 0.693 & 0.400       \\
                               & $w_2$ & 0.891             & 0.838 & 0.200       \\
                               & $w_3$ & 0.132             & 0.482 & 0.400       \\
        \midrule
        \multirow{3}{*}{PDPTW} & $w_1$ & 0.464             & 0.240 & 0.400       \\
                               & $w_2$ & 0.243             & 0.417 & 0.200       \\
                               & $w_3$ & 0.502             & 0.657 & 0.400       \\
        \midrule
        \multirow{3}{*}{CSP}   & $w_1$ & 0.362             & 0.239 & 0.400       \\
                               & $w_2$ & 0.059             & 0.577 & 0.200       \\
                               & $w_3$ & 0.770             & 0.195 & 0.400       \\
        \bottomrule
    \end{tabular}
    \egroup
    \normalsize
\end{table}

\begin{table}[t]
    \caption{PPO parameters.}
    \label{tab:parameters-ppo}
    \small
    \setlength\tabcolsep{4.5pt}
    \centering
    \begin{tabular}{lrrrrr}
        \toprule
        \multirow{2}{*}{Parameter}         & \multicolumn{2}{c}{TSP} & \multicolumn{2}{c}{PDPTW} & CSP                   \\
                                           & $R_2$                   & $R_3$                     & $R_2$ & $R_3$ & $R_1$ \\
        \midrule
        $c_1$ start                        & 0.24                    & 0.61                      & 0.09  & 0.53  & 0.34  \\
        $c_1$ end                          & 0.96                    & 0.04                      & 0.53  & 0.42  & 0.06  \\
        $c_1$ annealing                    & 196s                    & 56s                       & 41s   & 236s  & 196s  \\
        $c_2$ start                        & 0.05                    & 0.9                       & 0.27  & 0.19  & 0.39  \\
        $c_2$ end                          & 0.04                    & 0.88                      & 0.06  & 0.04  & 0.20  \\
        $c_2$ annealing                    & 171s                    & 241s                      & 171s  & 181s  & 431s  \\
        batch size                         & 345                     & 25                        & 725   & 995   & 1015  \\
        minibatch size                     & 5                       & 5                         & 255   & 445   & 15    \\
        \# epochs                          & 71                      & 65                        & 75    & 82    & 65    \\
        Clipping                           & 13.85                   & 44.09                     & 47.1  & 41.55 & 11.8  \\
        $\alpha$ actor ($\times 10^{-4}$)  & 0.176                   & 1.82                      & 35.9  & 0.175 & 18.1  \\
        $\alpha$ critic ($\times 10^{-4}$) & 0.826                   & 6.75                      & 0.514 & 22.8  & 7.31  \\
        \bottomrule
    \end{tabular}
    \normalsize
\end{table}

\begin{table}[t]
    \caption{DDQN parameters.}
    \label{tab:parameters-ddqn}
    \small
    \setlength\tabcolsep{4.5pt}
    \centering
    \begin{tabular}{lrrrrr}
        \toprule
        \multirow{2}{*}{Parameter}    & \multicolumn{2}{c}{TSP}   & \multicolumn{2}{c}{PDPTW} & CSP                                                                               \\
                                      & \multicolumn{1}{c}{$R_2$} & \multicolumn{1}{c}{$R_3$} & \multicolumn{1}{c}{$R_2$} & \multicolumn{1}{c}{$R_3$} & \multicolumn{1}{c}{$R_1$} \\
        \midrule
        $\alpha$ ($\times 10^{-2}$)   & 0.01                      & 0.01                      & 0.01                      & 0.2                       & 0.05                      \\
        $\epsilon_{DDQN}$             & 0.83                      & 0.65                      & 0.03                      & 0.38                      & 0.2                       \\
        Batch size                    & 19                        & 16                        & 236                       & 17                        & 250                       \\
        Gradient clipping             & 49.5                      & 45.5                      & 0.55                      & 30                        & 35                        \\
        Memory size ($\times 10^{3}$) & 5                         & 5                         & 10                        & 10                        & 10                        \\
        \bottomrule
    \end{tabular}
    \normalsize
\end{table}

\begin{table}[!ht]
    \caption{$\epsilon$-greedy and UCB parameters.}
    \label{tab:paramters-egreedy-ucb}
    \small
    \setlength\tabcolsep{3.5pt}
    \centering
    \begin{tabular}{llrrrrr}
        \toprule
        \multirow{2}{*}{Agent}             & \multirow{2}{*}{Param.} & \multicolumn{2}{c}{TSP}   & \multicolumn{2}{c}{PDPTW} & CSP                                                                               \\
                                           &                         & \multicolumn{1}{c}{$R_2$} & \multicolumn{1}{c}{$R_3$} & \multicolumn{1}{c}{$R_2$} & \multicolumn{1}{c}{$R_3$} & \multicolumn{1}{c}{$R_1$} \\
        \midrule
        \multirow{2}{*}{$\epsilon$-greedy} & $\epsilon$              & 0.015                     & 0.043                     & 0.006                     & 0.005                     & 0.405                     \\
                                           & $\alpha$                & 0.040                     & 0.581                     & 0.506                     & 0.523                     & 0.829                     \\
        \multirow{2}{*}{UCB}               & $c$                     & 0.332                     & 0.303                     & 0.395                     & 4.563                     & 4.281                     \\
                                           & $\alpha$                & 0.136                     & 0.307                     & 0.967                     & 0.618                     & 0.138                     \\
        \bottomrule
    \end{tabular}
\end{table}

\end{document}